\documentclass[sigconf]{acmart}
\copyrightyear{2025}
\acmYear{2025}
\setcopyright{acmlicensed}
\acmConference[KDD '25] {Proceedings of the 31st ACM SIGKDD Conference on Knowledge Discovery and Data Mining V.2}{August 3--7, 2025}{Toronto, ON, Canada.}
\acmBooktitle{Proceedings of the 31st ACM SIGKDD Conference on Knowledge Discovery and Data Mining V.2 (KDD '25), August 3--7, 2025, Toronto, ON, Canada}
\acmISBN{979-8-4007-1454-2/25/08}
\acmDOI{10.1145/3711896.3737197}
\usepackage{bm}
\usepackage{amsthm}
\usepackage{mathtools}
\usepackage{enumitem} % for nice itemize
\usepackage{multirow}
\usepackage{subfigure}
\usepackage{stfloats} % stick table top in the current page

\theoremstyle{plain}
\newtheorem{theorem}{Theorem}

\newtheorem{definition}[theorem]{Definition}

\theoremstyle{remark}

\usepackage{xspace}
\newcommand{\paratitle}[1]{\vspace{1.2ex}\noindent\textbf{#1}}
\newcommand{\fig}{Fig.\xspace}
\newcommand{\tab}{Tab.\xspace}

\newcommand{\equ}{Eq.\xspace}

\newcommand{\ie}{\emph{i.e., }}

\newcommand{\eg}{\emph{e.g., }}

\newcommand{\etc}{\emph{etc}}
\newcommand{\wrt}{\emph{w.r.t. }}
\newcommand{\model}{Bid2X\xspace}

\begin{document}

% \title{Bid2X: A Transformer-based Foundation Model for Auto-bidding via Multivariate Time Series Modeling}
% \title{Revealing Dynamics of Ad Bidding Environment:\\A Foundation Model Lens}
\title{Bid2X: Revealing Dynamics of Bidding Environment in Online Advertising from A Foundation Model Lens}

\iffalse
% \affiliation{%
\author{Jiahao Ji}
\authornote{This work was done during the internship at Alibaba Group.}
\orcid{0000-0003-3029-2262}
\affiliation{%
  \institution{School of Computer\\Science and Engineering,\\Beihang University}
  \city{Beijing}
  \country{China}
}

\author{Tianyu Wang}
\orcid{0009-0006-0292-2078}
\author{Yeshu Li}
\orcid{0000-0001-5075-1062}
\author{Yusen Huo}
\orcid{0009-0006-8863-3209}
\author{Zhilin Zhang}
\orcid{0000-0001-6665-2288}
\affiliation{%
  \institution{Alibaba Group}
  \city{Beijing}
  \country{China}
}

\author{Chuan Yu}
\author{Jian Xu}
\authornote{Corresponding author: xiyu.xj@alibaba-inc.com.}
\orcid{0000-0003-3111-1005}
\author{Bo Zheng}
\orcid{0000-0002-4037-6315}
\affiliation{%
  \institution{Alibaba Group}
  \city{Beijing}
  \country{China}
}
\fi 

% \iffalse
\author{Jiahao Ji}
\authornote{This work was done during the internship at Alibaba Group.}
\orcid{0000-0003-3029-2262}
% \affiliation{%
%   \institution{Beihang University}
%   \city{Beijing}
%   \country{China}
% }
\affiliation{%
  \institution{Alibaba Group}
  \city{Beijing}
  \country{China}
}
\author{Tianyu Wang}
\orcid{0009-0006-0292-2078}
\affiliation{%
  \institution{Alibaba Group}
  \city{Beijing}
  \country{China}
}
\author{Yeshu Li}
\orcid{0000-0001-5075-1062}
\affiliation{%
  \institution{Alibaba Group}
  \city{Beijing}
  \country{China}
}
\author{Yusen Huo}
\orcid{0009-0006-8863-3209}
\affiliation{%
  \institution{Alibaba Group}
  \city{Beijing}
  \country{China}
}
\author{Zhilin Zhang}
\orcid{0000-0001-6665-2288}
\affiliation{%
  \institution{Alibaba Group}
  \city{Beijing}
  \country{China}
}
\author{Chuan Yu}
\affiliation{%
  \institution{Alibaba Group}
  \city{Beijing}
  \country{China}
}
\author{Jian Xu}
\authornote{Corresponding author: xiyu.xj@alibaba-inc.com.}
\orcid{0000-0003-3111-1005}
\affiliation{%
  \institution{Alibaba Group}
  \city{Beijing}
  \country{China}
}
\author{Bo Zheng}
\orcid{0000-0002-4037-6315}
\affiliation{%
  \institution{Alibaba Group}
  \city{Beijing}
  \country{China}
}
% \fi

% \renewcommand{\shortauthors}{Ji et al.}
\renewcommand{\shortauthors}{Jiahao Ji et al.}

\begin{abstract}
Auto-bidding is crucial in facilitating online advertising by automatically providing bids for advertisers.
While previous work has made great efforts to model bidding environments for better ad performance, it has limitations in generalizability across environments since these models are typically tailored for specific bidding scenarios. 
To this end, we approach the scenario-independent principles through a unified function that estimates the achieved effect under specific bids, such as budget consumption, gross merchandise volume (GMV), page views, etc. Then, we propose a bidding foundation model Bid2X to learn this fundamental function from data in various scenarios.
Our Bid2X is built over uniform series embeddings that encode heterogeneous data through tailored embedding methods.
To capture complex inter-variable and dynamic temporal dependencies in bidding data, we propose two attention mechanisms separately treating embeddings of different variables and embeddings at different times as attention tokens for representation learning.
On top of the learned variable and temporal representations, a variable-aware fusion module is used to perform adaptive bidding outcome prediction. 
To model the unique bidding data distribution, we devise a zero-inflated projection module to incorporate the estimated non-zero probability into its value prediction, which makes up a joint optimization objective containing classification and regression. The objective is proven to converge to the zero-inflated distribution. 

Our model has been deployed on the ad platform in Taobao, one of the world's largest e-commerce platforms. Offline evaluation on eight large-scale real-world datasets exhibits Bid2X's superiority compared to various baselines and its generality across different scenarios. In real-world applications, Bid2X increased GMV by 4.65\% and ROI by 2.44\% in online A/B tests, paving the way for the bidding foundation model in computational advertising.
\end{abstract}



% \keywords{Online advertising, Auto-bidding, Foundation model}


\maketitle
% 总共8页
% ================Main body starts here====================
\section{Introduction}

% With the increasing digitization of business, online advertising platforms have become indispensable tools for advertisers to target audiences and drive sales effectively. 
With the increasing automation of online advertising platforms, auto-bidding services (\ie automatically adjust bids for each ad impression) have become indispensable tools for advertisers to achieve ad performance goals in various advertising scenarios~\cite{wang2017display,zhang2012joint,ren2017bidding,lin2016combining}. 
On designing auto-bidding algorithms for e-commerce advertising, it is crucial to have a thorough understanding of the current bidding environment~\cite{guo2024aigb,ou2023survey}. 
Specifically, for an ad campaign, given a specific bid at a certain time slot, how much revenue will the advertiser earn and what is the total cost to pay for the revenue? 
Capturing such dynamic bidding environments can aid in responding more timely and accurately to changes in the marketplace, resulting in better ad performance~\cite{wen2022cooperative}.

\begin{figure}[t]
    \centering
    \includegraphics[width=0.93\columnwidth]{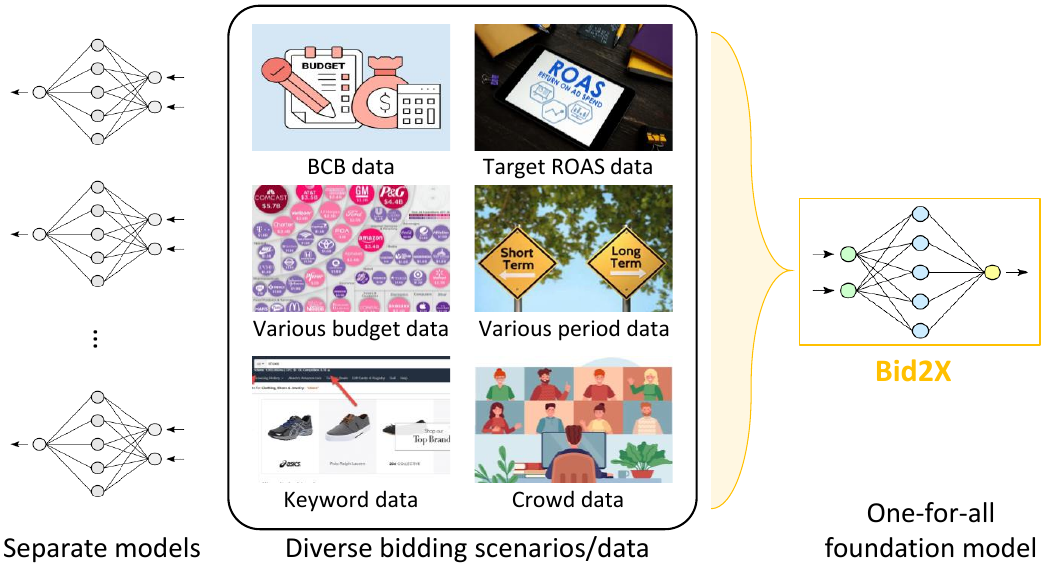}%\vspace{-2mm}
    % \caption{Existing separate models v.s. a one-for-all foundation model for bidding environment modeling.}
    \caption{Existing models are designed and trained for specific bidding scenarios, so they cannot adapt to various scenarios. In contrast, we propose to train a one-for-all foundation model on broad bidding data, enabling it to handle a wide range of bidding scenarios.}
    \label{fig:motivation}%\vspace{-4mm}
\end{figure}

Existing auto-bidding algorithms usually model the bidding environment implicitly, having limitations in comprehensively understanding the bidding environment and generalizing across environments. 
For example, linear programming (LP) computes the maximum total value of the impressions obtainable for a given bid based on historical bidding environments~\cite{zhu2017optimized,zhou2008budget,grigas2017profit,he2021unified}.
Proportional–Integral–Derivative (PID)~\cite{doyle2013feedback} based methods portray the possible cost for a given bid~\cite{moriwaki2021real,grislain2019recurrent,zhang2016feedback}. Reinforcement learning-based methods typically employ an environment model to learn the relation between the bid and cost or reward~\cite{cai2017real,mou2022sustainable}.
These environment modeling processes can be summarized as a unified environment function $f_\theta(Y|bid_t)$, where $Y$ involves target variables such as cost, reward, and count of the winning impressions, and $t$ denotes the timestamp. 
% More importantly, this ``bidding environment modeling'' task can be used as a fundamental task for auto-bidding.
% Although these 
Although these approaches depict bidding environments from different perspectives, none of them can provide a comprehensive description of the bidding environments.
% 各场景单独建模（关键词，人群，智能...），不同出价类型（MaxReturn，TargetCPX，TargetRoas...）分别建模，如果加上短视频、直播等，至少有20个不同类型的出价模型在线上服务。每个模型得单独优化和维护，还是需要花费较多的时间和资源。现在大家的精力只能够维护几个重点模型。我们需要建设一个独立于具体场景和计划类的的竞价环境模型，并通过显式或隐式接入到现有的出价模型中，提升效果&降低维护成本。
Moreover, as in \fig~\ref{fig:motivation}, existing models are usually designed and trained for specific scenarios~\cite{zhou2008budget,moriwaki2021real,cai2017real}, lacking adaptability to new bidding scenarios and versatility across scenarios.
% limiting their capability in accommodating different scenarios with dynamic bidding environments.
% lacking flexibility and versatility to adapt to new bidding environments.
%todo BFM Bid2X?

To overcome these problems, we propose to train a bidding foundation model, which is supposed to be ``\textit{a large deep learning model trained on broad bidding environment data such that it can be applied across a wide range of bidding scenarios}''. The bidding foundation model is versatile across various scenarios due to the existence of generalized principles that are independent of specific bidding environments. For example, more cost-effective impressions bring better ad performance, there exist closeness and periodicity in bidding environments, and the bid and its outcome follow the law of diminishing marginal returns.
% 出价与价值和扣费成正相关关系，和该时段用户的流量性成负相关关系。
% 出价和竞得价值的关系存在边际效益递减现象
% 流量具有连续性，捕捉到了时序上的相关性，大促流量分布漂移时，适应性输出花费

However, it remains open to both academia and industry on how to build a bidding foundation model for industrial e-commerce advertising. We consider three challenges in this direction. 
(1) \textbf{Heterogeneous bidding data}. Data collected from different bidding scenarios lack uniformity and primarily consist of point data without temporal information, time series data, as well as discrete data and continuous data. Each type of data has its own characteristics, and there is currently no unified method to encode all of them.
(2) \textbf{Complex and dynamic data dependency}. The real-world bidding environment is a highly dynamic multi-intelligence game process, which leads to complex dependencies between the control variable (bid) and target variables (cost, reward, count) in bidding data. Furthermore, these dependencies evolve with time, \eg the same bid may bring more rewards at night than during working hours because people are more likely to shop online after work. Existing methods mainly learn dependencies between variables but fail to consider temporal dynamics.
(3) \textbf{Unique data distribution}. As engaging in the bidding process does not guarantee winning impressions, the bidding data often include many zero values, resulting in a zero-inflated distribution~\cite{lambert1992zero}. This unique distribution violates the normal data distribution assumption of existing neural network models, resulting in sub-optimal performance.

% Bidding data has a unique distribution called the zero-inflated distribution, which allows for frequent zero-valued observations. This is caused by the fact that participating in the bidding does not necessarily mean winning for impressions, so the data distribution will contain a large number of zero values. This violates the normal distribution assumption in the common optimization objective for existing neural network-based models, resulting in sub-optimal performance.

To address these challenges, we propose a bidding foundation model coined \model, which leverages a conditional next-token prediction approach to learn universal principles from diverse bidding data across scenarios.
\textbf{First}, we process different types of bidding data to encode them as uniform series representations, which provides input for dependency modeling.
\textbf{Second}, to model complex and dynamic dependencies, we apply attention mechanisms to variable and temporal dimensions to learn the correlations from both perspectives. Then, we design a variable-aware fusion module to refine the inter-variable dependencies (learned from historical bidding data) with the bidding series of the day of interest. This provides a more accurate reference for the day's bidding outcome prediction and improves our model's adaptability to the day's dynamic dependencies.
\textbf{Third}, to model the unique distribution of bidding data, we devise an end-to-end zero-inflated projection module that combines the model's prediction of the probability of a non-zero value and the magnitude of that value into a joint optimization objective. Moreover, we theoretically demonstrate that this objective can converge to the real distribution of bidding data under mild assumptions. Our contributions are four-fold:
\begin{itemize}[leftmargin=*]
    \item We introduce the novel concept of the bidding foundation model for bidding environment modeling, marking a significant innovation in the field of computational advertising. This paradigm innovation moves beyond the limitations of traditional, scenario-specific models and provides a solution capable of generalizing across various bidding environments.
    \item We for the first time identify three unique challenges of the bidding environment modeling problem, which are fundamental to developing bidding foundation models and advancing the capability of computational advertising systems.
    \item We propose a bidding foundation model \model that can unify heterogeneous bidding data as series embedding and learn complex and dynamic dependencies to achieve generalization across environments. We theoretically ensure that the model can converge to the zero-inflated data distribution.
    \item Extensive experiments on eight real-world datasets demonstrate the superior generalization of \model in various scenarios compared to several baselines. Online results in Taobao, a large e-commerce company, further verify our model's effectiveness.
\end{itemize}
% \begin{itemize}[leftmargin=*]
%     \item We summarize for the first time the cornerstone task of auto-bidding, \ie bidding environment modeling, and give the definition and mechanism of the BFM.
%     \item We introduce a foundation model \model that can unify heterogeneous bidding data as series embedding and learn complex and dynamic bidding environments to achieve generalization across environments. We theoretically ensure that the model can converge to the zero-inflated distribution of bidding data.
%     %todo which effective?
%     \item Extensive experiments on eight real-world datasets demonstrate the superior generalization of \model in various scenarios compared to seven baselines. Online results on Taobao, a large e-commerce company, further verify our model's effectiveness.
% \end{itemize}

\section{Related Work}

\paratitle{Auto-bidding and Bidding Environment Modeling.} 
Auto-bidding techniques in computational advertising have been a research focus in recent years~\cite{zhang2012joint,ren2017bidding,maehara2018optimal}. 
They are employed to automatically place bids on ad spaces, offering convenience and efficiency in managing bids compared to manual bidding. 
The main focus of such techniques is to optimize given key performance indicators (\eg clicks or conversions) while maintaining a certain budget ~\cite{guo2024aigb,wang2017display}. 
A straightforward solution for auto-bidding is linear programming based on the historical bidding environment~\cite{lin2016combining,zhou2008budget, zhu2017optimized,grigas2017profit,he2021unified}. While it is highly efficient and simple to implement, it does not take the dynamic nature of the bidding environment into account.
To model such dynamics, some work utilizes the PID algorithm from the control theory~\cite{doyle2013feedback} to adjust the bidding strategy based on the latest observed data. Some typical examples are shown in~\cite{grislain2019recurrent,zhang2016feedback,moriwaki2021real}, which aim to control different variables such as cost-per-click, budget pacing, \etc. 
These controlling methods enable the bidding agents to perceive the current environment in real time and then update the bids that were once based on outdated data~\cite{ou2023survey}.
Recently, reinforcement learning (RL) based methods have emerged as a promising approach due to their ability to effectively handle the sequential nature of the bidding process~\cite{zhao2019deep, he2021unified, hu2025learning}. 
For example, Cai et al.~\cite{cai2017real} designed a bidding environment and applied a deep RL algorithm to learn the optimal bidding strategy. Zhang et al.~\cite{mou2022sustainable} adopted the RL framework for auto-bidding and showed that their approach can outperform traditional bidding strategies. 
% Besides, landscape forecasting approaches~\cite{ren2019deep, li2022arbitrary} that forecast the probability distribution of market price for each ad auction can also be viewed as bidding environment modeling methods, but they do not directly predict the bidding outcomes.
In addition, landscape forecasting methods~\cite{ren2019deep, li2022arbitrary} that predict the probability distribution of market prices for each advertising auction can also be regarded as bidding environment modeling methods, but they do not directly predict bidding results.
% Wen et al.~\cite{wen2022cooperative} even proposed a multi-agent-based approach that enables the modeling of multiple auto-bidding agents simultaneously to include more environmental information.
However, these methods are usually designed for specific scenarios and cannot be well generalized across various bidding scenarios.

% mainly focuses on the mechanism design in display advertising RTB from the platform's perspective, whereas our paper addresses the fundamental issue of bidding environment modeling from the advertiser's perspective.

\paratitle{Foundation Model.} Foundation models are general-purpose technologies that can support a diverse range of scenarios~\cite{bommasani2021opportunities}. Typically, they are large deep-learning models trained on broad and massive data.
They have been developed across a range of modalities, such as text~\cite{zhao2023survey}, vision~\cite{awais2023foundational}, graph~\cite{liu2023towards}, time series~\cite{liang2024foundation}, and even robotics~\cite{firoozi2023foundation} and autonomous driving~\cite{chen2024end}. However, the foundation model in computational advertising has rarely been explored. 

\paratitle{Time Series Prediction.} Time series prediction (TSP) aims to fit a model using historical (time-stamped) data to predict future values of variables. It is crucial in fields like epidemiology~\cite{wang2023high}, transportation~\cite{ji2025seeing,ji2023modeling,ji2022stden,wang2022traffic,ji2020interpretable}, and meteorology~\cite{hettige2024airphynet}. 
Extensive research has been conducted in both traditional statistics~\cite{box1968some,holt2004forecasting} and deep learning~\cite{schmidhuber1997long,rangapuram2018deep}. 
Recent studies have seen a surge in Transformer-based TSP models~\cite{liuitransformer,wen2023transformers,zhang2023crossformer,zhou2022fedformer,wu2021autoformer,zhou2021informer}, which can be categorized into two groups: temporal-wise and variable-wise.
The former builds input tokens from series values at each time slot, which has been widely researched~\cite{zhang2023crossformer,zhou2021informer}.
However, these studies are confronted with issues like timestep mixing and permutation invariance, leading to underperformance on some datasets compared to simple linear predictors~\cite{liuitransformer,zeng2023transformers}.
To tackle this, variable-wise Transformers that consider series data of each variable as a token are proposed and show the state-of-the-art performance in TSP. 
However, these methods have not yet been explored in the bidding environment modeling problem and we make the first attempt.  
% formalize the bidding environment modeling as a conditional time series prediction problem and 

\section{Preliminaries}
In this section, we first define some basic concepts and then introduce the studied problem of this paper. 

% \subsection{Basic Concept}

\begin{definition}[Ad Campaign]
An ad campaign $\mathcal{C}$ is an order in which an advertiser seeks product promotion constrained by budget, product category, advertiser category, and delivery start/end times. It also involves some contextual information such as total historical clicks, total historical costs, total historical cost effectiveness, \etc.
\end{definition}

In online advertising platforms, an ad campaign is usually settled and reset at the end of each day. Therefore, we denote the $i$-th ad campaign at day $\tau$ as {\small$\mathcal{C}_i^{(\tau)}$}.

\begin{definition}[Bidding Trajectory]
A bidding trajectory is a sequence of bidding records generated in the execution process of an ad campaign {\small$\mathcal{C}_i^{(\tau)}$}, denoted as {\small$\bm{X}_{i}^{(\tau)} = (\bm{x}_{i, 1}^{(\tau)}, \dots, \bm{x}_{i, m}^{(\tau)})$}, where $m$ is the number of bidding records. Each record $\bm{x} = (b, c, r, ct, t)$, where $b$ is the bid, $c, r$, and $ct$ represent the cumulative cost, reward, and count of winning impressions over the time from timestamp $t$ to the time when the bid is adjusted respectively. We call a bidding trajectory \textbf{complete} when the corresponding ad campaign is finished, otherwise it is \textbf{incomplete}.
\end{definition}

Note that in a bidding record, $b$ is the optimal bidding parameter introduced in \cite{he2021unified}, which can be transformed into the real bid. 
In general, one day is divided into several time slots (ticks) according to the minimum bid adjustment period. Accordingly, we can map each bid record into a time slot. 
We observe that some time slots have no bidding record, and a bidding trajectory does not necessarily start from the first time slot and end at the last time slot.

% \subsection{Problem Formulation}

\begin{definition}[Bidding Environment Modeling Problem]\label{def:problem}
Given historical bidding data {\small$\{\bm{X}_i^{(\tau-1)}, \mathcal{C}_i^{(\tau-1)} \}$}, today's bidding data until time slot $t$, \ie {\small$\{\bm{X}_i^{(\tau)}, \mathcal{C}_i^{(\tau)}\}$}, and bid information {\small$b_{i,t+1}^{(\tau)}$} of the next time slot $t+1$, the \textbf{bidding environment modeling} problem aims to predict the corresponding outcomes via a function $f_{\Theta}(\cdot)$:
\begin{equation}\small
    \bm{y}_{i,t+1}^{(\tau)} = f_{\Theta}\left(b_{i,t+1}^{(\tau)}, \bm{X}_i^{(\tau)}, \mathcal{C}_i^{(\tau)}, \bm{X}_i^{(\tau-1)}, \mathcal{C}_i^{(\tau-1)}\right),
\end{equation}
where the output includes the cost, reward, and count at the $(t+1)$-th time slot, \ie {\small$\bm{y}_{i,t+1}^{(\tau)} = (c_{i,t+1}^{(\tau)}, r_{i,t+1}^{(\tau)}, ct_{i,t+1}^{(\tau)})$}. 
\end{definition}

We leverage the temporal nature of the data and frame the bidding environment modeling problem as a self-supervised task that predicts the most likely outcome at the next time step (given a bid) based on previous bidding records, \ie conditional next-token prediction. Our problem formulation does not require manual annotations, as both the input and target are inherently present in the data, akin to how they are in language data. This novel self-supervised problem formulation approach facilitates effectively capturing the temporal dynamics of the bidding environment and paves the way for foundation models in this domain.

It is worth noting that in Definition~\ref{def:problem} the historical bidding trajectories are complete whereas today's bidding trajectories are still in progress. Therefore, we utilize the historical trajectory to learn the complex correlations among different variables (\eg bid and cost) and today's incomplete trajectory to capture the dynamic temporal dependency.
To avoid clutter of notations, \textit{we omit the sample index $i$ for all symbols} in the rest of this paper.

% This problem is also a conditional next-token prediction task.
% It is worth noting that the bidding trajectory in historical data is complete whereas today's bidding trajectories are still in progress. Therefore, we utilize the historical trajectory to learn the complex correlations among different variables (\eg bid and cost) and today's incomplete trajectory to capture the dynamic temporal dependency.
% To avoid clutter of notation, \textit{we omit the sample index $i$ for all symbols} in the rest of this paper.

% \paratitle{Remark:} We leverage the temporal nature of the data and frame the bidding environment modeling problem as a self-supervised task that predicts the most likely outcome at the next time step (given a bid) based on previous bidding records. Our problem formulation does not require manual annotations, as both the input and target are inherently present in the data, akin to how they are in language datasets. This novel self-supervised problem formulation approach facilitates effectively capturing the temporal dynamics of the bidding environment and paves the way for foundation models in this domain.

\begin{figure}[t]
    \centering
    \includegraphics[width=0.93\columnwidth]{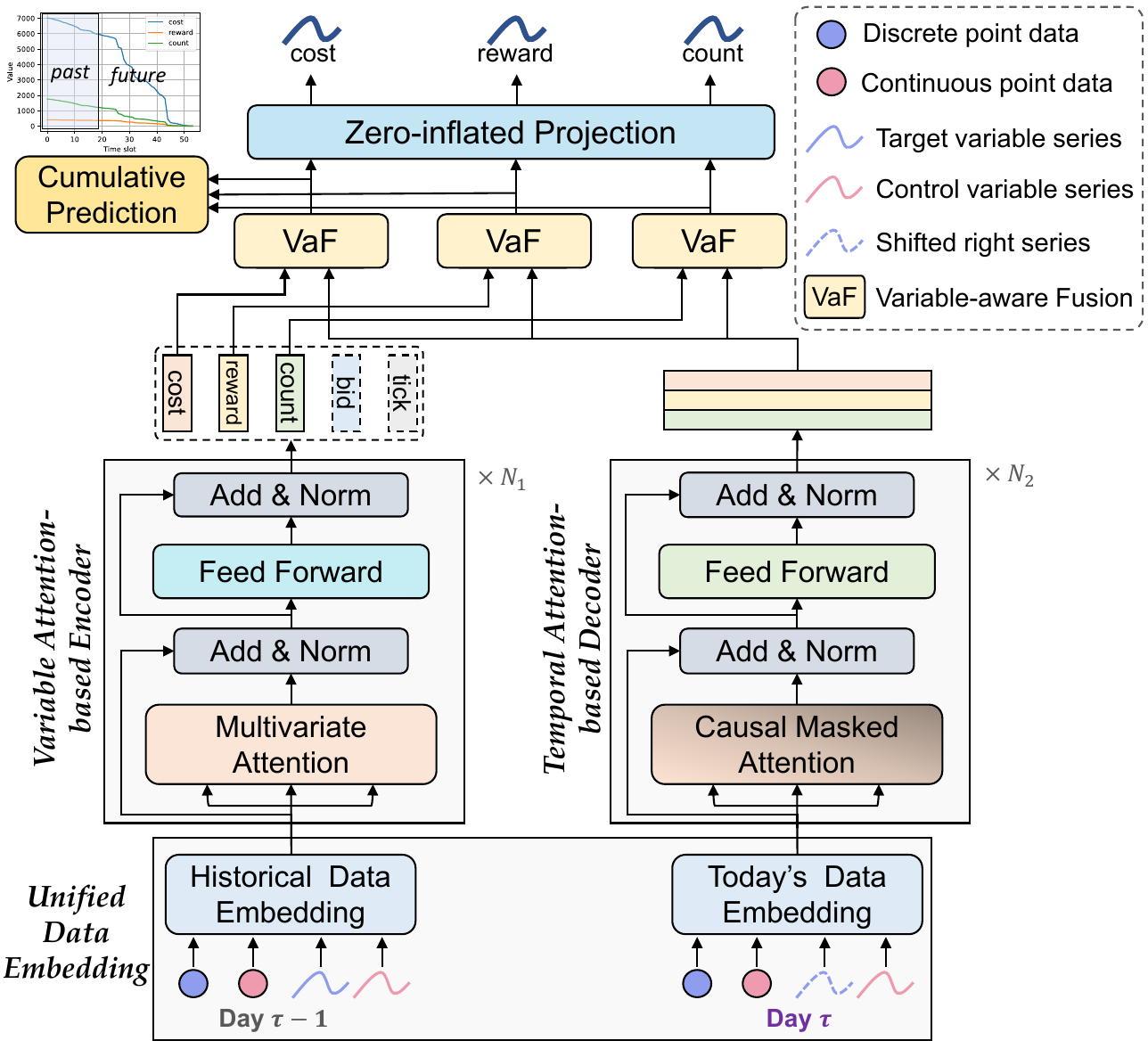}%\vspace{-2mm}
    \caption{The overall architecture of our \model model.}%\vspace{-4mm}
    \label{fig:framework}
\end{figure}
% \begin{figure*}[ht]
%     \centering
%     \includegraphics[width=0.85\linewidth]{figures/framework.pdf}
%     \vspace{-.2cm}\caption{The overall architecture of our bidding foundation model: \model.}
%     \label{fig:framework}\vspace{-.4cm}
% \end{figure*}

\section{Method}\label{sec:method}

In this section, we give a detailed introduction of the proposed \model foundation model with illustration in \fig~\ref{fig:framework}. 
First, we provide a unified data embedding method for heterogeneous bidding data in Section~\ref{sec:unified_data_emb}. 
Then, we elaborate on the bidding transformer used for variable and temporal dependency modeling in Section~\ref{sec:dep_model}.
Finally, we detail a zero-inflated projection for the unique distribution of bidding data, and introduce a self-supervised auxiliary task to complement information from a global view in Section~\ref{sec:zero_inflated}.

\subsection{Unified Data Embedding}\label{sec:unified_data_emb}

This part aims to transform heterogeneous bidding data into a unified series embedding with tailored embedding methods. Since historical data and today's data are used for modeling different types of dependencies, we embed them via separate modules.  

\paratitle{Historical Data Embedding.} 
Historical data are utilized for inter-variable correlation modeling, so we propose to transform each variable series into an independent embedding. 

Specifically, given the cost series {\small$\bm{c}^{(\tau-1)} = (c_{1}^{(\tau-1)}, \dots, c_{T_{\tau-1}}^{(\tau-1)})$} of trajectory {\small$\bm{X}^{(\tau-1)}$} with length {\small$T_{\tau-1}$}, we encode it as {\small$\bm{z}_{c}^{(\tau-1)} \in \mathbb{R}^{D}$} via
\begin{equation}\small
    \bm{z}_{c}^{(\tau-1)} = \bm{c}^{(\tau-1)} \cdot \bm{W}_c^\top,
\end{equation}
where {\small$\bm{W}_c \in \mathbb{R}^{D \times T_{\tau-1}}$} is learnable parameters. Since different bidding trajectories have different lengths, we pad all trajectories to the maximum length $T$. The shape of {\small$\bm{W}_c$} is therefore $D \times T$.
Similar to the cost series, we can generate embeddings of other variables. By using these embeddings, we can form an embedding matrix {\small$\bar{\bm{Z}}^{(\tau-1)}_s \in \mathbb{R}^{C_0 \times D}$}, where $C_0$ is the number of input variables.

Besides the series data, there is some constraint and context information in ad campaign data {\small$\mathcal{C}^{(\tau-1)}$}, including discrete and continuous data. 
For the discrete one-hot categorical constraints such as the advertiser category, an embedding layer is used to transform them into low-dimensional dense representations. 
For the $j$-th categorical constraint {\small$\bm{t}_j$, let $\bm{E}^{(j)} = (\bm{e}_1^{(j)}, \dots, \bm{e}_{K_j}^{(j)})^\top \in \mathbb{R}^{K_j \times D}$} represent the $j$-th embedding dictionary, where {\small$\bm{e}^{(j)}_k \in \mathbb{R}^{D}$} is an embedding vector with dimensionality of $D$. The embedding operation follows the table lookup mechanism, \ie if the $i$-th entry of $\bm{t}_j$ is $t_{j,i} = 1$, the embedded representation of $\bm{t}_j$ is a single embedding vector $\bm{e}_i^{(j)}$.
For the continuous numeric constraints like budget, we separately feed each constraint into different linear transformations to generate embedding. The same transformation is employed for context data as they are all numeric features. 
Next, all ad campaign embeddings are summed as a single context embedding {\small$\bar{\bm{Z}}_c^{(\tau-1)} \in \mathbb{R}^{D}$}.

Finally, we sum up the variable embedding {\small$\bar{\bm{Z}}_s^{(\tau-1)}$} and the context embedding {\small$\bar{\bm{Z}}_c^{(\tau-1)}$} to produce the historical data embedding via
\begin{equation}\small
    \bm{Z}^{(\tau-1)} = \bar{\bm{Z}}_s^{(\tau-1)} \oplus \bar{\bm{Z}}_c^{(\tau-1)},
\end{equation}
where $\oplus$ is the broadcasting addition operation, which first duplicates the lower-dimensional matrix to match the shape of the higher-dimensional matrix and then adds them together.  

% Each variable embedding will be considered as a token to model inter-variable correlation in Section~\ref{sec:dep_model}. 

\paratitle{Today's Data Embedding.} 
Today's data are utilized for temporal dependency modeling, so we treat all values of a time slot as a token and encode them as a $D$-dimensional embedding. 

Before delving into the embedding layer, we preprocess the bidding trajectory to avoid information leakage and facilitate modeling convenience.
Specifically, given bidding trajectory until the current time slot $t$: {\small$\bm{X}^{(\tau)} = (\bm{x}_{1}^{(\tau)}, \dots, \bm{x}_{t}^{(\tau)}) \in \mathbb{R}^{C_0 \times t}$}, we split it into two parts along the variable dimension $C_0$: {\small$\bm{X}_p^{(\tau)} \in \mathbb{R}^{C_1 \times t}$} that contains control variables (bid and tick time) and {\small$\bm{X}_q^{(\tau)} \in \mathbb{R}^{C_2 \times t}$} that contains target variables (cost, reward, and count).
To avoid information leakage, we shift right the target series and use a zero vector as the start token, making {\small$\bm{X}_p^{(\tau)} = (\bm{0}, \bm{x}_{p, 1}^{(\tau)}, \dots, \bm{x}_{p, t}^{(\tau)})$}. 
Meanwhile, we include the future bid information at the end of the control series, making {\small$\bm{X}_q^{(\tau)} = (\bm{x}_{q, 1}^{(\tau)}, \dots, \bm{x}_{q, t}^{(\tau)}, \bm{x}_{q, t+1}^{(\tau)})$}.
Then, we stack these two series back to the original format and obtain {\small$\bm{X}^{(\tau)} \in \mathbb{R}^{C_0 \times (t+1)}$}. 
It is further padded to the max length $T$ along the time dimension $t+1$ for computational convenience because different bidding trajectories have different lengths, resulting in the input shape $C_0 \times T$.

Based on the preprocessed bidding trajectory {\small$\bm{X}^{(\tau)} \in \mathbb{R}^{C_0 \times T}$}, we transform the $t'$-th token {\small$\bm{x}_{t'}^{(\tau)} \in \mathbb{R}^{C_0}$} as
\begin{equation}\small
     \bm{z}_{t'}^{(\tau)} = \bm{x}_{t'}^{(\tau)} \cdot \bm{W}_x^\top,
\end{equation}
where {\small$\bm{W}_x \in \mathbb{R}^{D \times C_0}$}. 
By applying the transformation to all time slots, we can obtain an embedding {\small$\bar{\bm{Z}}_s^{(\tau)} \in \mathbb{R}^{T \times D}$}.
This embedding involves the global timestamp by a learnable time embedding in control variables, but the local positional information of the tokens in the sequence is not preserved. 
To this end, we add positional encoding to {\small$\bar{\bm{Z}}_s^{(\tau)}$} with a fixed position embedding matrix $\bm{P}$ defined by {\small$p_{i, 2d} = \sin(i/(2T)^{2d/D})$ and $p_{i, 2d+1} = \cos(i/(2T)^{2d/D})$}.

By repeating the same process for campaign data {\small$\mathcal{C}^{(\tau)}$} as that in the historical data embedding module, we can have a context representation {\small$\bm{Z}_c^{(\tau)} \in \mathbb{R}^{D}$}. On top of this, we generate today's data embedding as follows:
\begin{equation}\small
    \bm{Z}^{(\tau)} = \bar{\bm{Z}}_s^{(\tau)} \oplus \bar{\bm{Z}}_c^{(\tau)}.
\end{equation}

\subsection{Bidding Transformer}\label{sec:dep_model}
% \subsection{Dependency Modeling}\label{sec:dep_model}

This part aims to inject the complex inter-variable correlation into historical embedding {\small$\bm{Z}^{(\tau-1)}$} and dynamic temporal dependency into today's embedding {\small$\bm{Z}^{(\tau)}$} with two attention mechanisms. 
Then, we fuse {\small$\bm{Z}^{(\tau-1)}$} and {\small$\bm{Z}^{(\tau)}$} for a comprehensive understanding of bidding environments by a variable-aware fusion module. 
% refine the dependency embedding $\bm{Z}^{(\tau)}$ by the variable correlation embedding $\bm{Z}^{(\tau-1)}$, 
% embeddings correlations embedding by the dependency embedding via a variable-aware fusion module.

% \subsubsection{\textbf{Variable and Temporal-based Attention}} 
% To jointly model bidding environments from different perspectives, we design variable-based attention and temporal-based attention as follows.

\subsubsection{\textbf{Variable Attention-based Encoder}} 
We first model the complex correlation between different variables (\eg bid and cost, bid and reward, \etc.) with a variable attention mechanism, where the model treats each variable as a token and learns pairwise variable relationships. 
Specifically, given historical embedding {\small$\bm{Z}^{(\tau-1)} \in \mathbb{R}^{C_0 \times D}$}, we map it into {\small$\bm{Q}^{(\tau-1)}, \bm{K}^{(\tau-1)}, \bm{V}^{(\tau-1)} \in \mathbb{R}^{C_0 \times D}$} by three linear projections. 
Let {\small$\bm{q}_m^{(\tau-1)}, \bm{k}_n^{(\tau-1)}$} stand for the $m$-th and $n$-th row in {\small$\bm{Q}^{(\tau-1)}, \bm{K}^{(\tau-1)}$} respectively. We can compute the correlation between the $m$-th and $n$-th variables by
\begin{equation}\small
    \alpha_{m, n} = \frac{\exp\left((\bm{q}_m^{(\tau-1)})^\top \cdot \bm{k}_n^{(\tau-1)} / \lambda\right)}{\sum_{n' = 1}^{C_0} \exp\left((\bm{q}_m^{(\tau-1)})^\top \cdot \bm{k}_{n'}^{(\tau-1)} / \lambda\right)},
\end{equation} 
where $\lambda$ is the scale factor and is set to $\sqrt{D}$. By calculating all $\alpha_{m, n}$, we can have a variable correlation map {\small$\bm{A} \in \mathbb{R}^{C_0 \times C_0}$} that exhibits the multivariate correlations between paired variables.
Consequently, highly correlated variables will be more weighted for the next representation interaction with {\small$\bm{V}^{(\tau-1)}$}. The interaction is formulated as {\small$\mathrm{LN}(\bm{Z}^{(\tau-1)} + \bm{A}\cdot\bm{V}^{(\tau-1)})$}, where {\small$\mathrm{LN}$} denote layer normalization. 
After that, all variables' representations are independently processed by a shared feed-forward network, which aims to portray the intrinsic properties of every variable, such as the amplitude and trend.
The above modules make up a variable-based attention block. By stacking {\small$N_1$} such blocks, we obtain the output representation {\small$\bm{H}^{(var)} \in \mathbb{R}^{C_0 \times D}$} that fully captures the inter-variable correlation.

% \paratitle{Temporal-based Attention.} 
\subsubsection{\textbf{Temporal Attention-based Decoder}} 
Apart from the inter-variable correlation, the temporal dependency evolving with time is also an important perspective of biding environments. We capture such dynamic dependency along the temporal dimension with a causal attention mechanism, where the model treats each time slot as a token and only attends to tokens in the past.
Specifically, given today's embedding {\small$\bm{Z}^{(\tau)} \in \mathbb{R}^{T \times D}$}, we generate queries, keys, and values as {\small$\bm{Q}^{(\tau)}, \bm{K}^{(\tau)}, \bm{V}^{(\tau)} \in \mathbb{R}^{T \times D}$} for attention computation. 
The temporal attention map {\small$\bm{B} \in \mathbb{R}^{T\times T}$} is then computed as
\begin{equation}\small
    \bm{B} = \mathrm{Softmax}\left(\frac{\bm{Q}^{(\tau)}\cdot (\bm{K}^{(\tau)})^\top}{\sqrt{D}} \odot \bm{M}\right),
\end{equation}
where {\small$\bm{M} = \bm{L}_{1} + \bm{U}_{-\inf} \in \mathbb{R}^{T \times T}$} is the causal masking matrix with {\small$\bm{L}_1$} a one-valued lower triangular matrix and {\small$\bm{U}_{-\inf}$} a $(-\inf)$-valued strict upper triangular matrix. 
The masking matrix leads to {\small$\bm{B}$} a lower triangular matrix in which all the entries above the main diagonal are zero.
This ensures no information leakage during temporal dependency learning because our model only attends to past information and cannot see future ones.
On top of the learned temporal attention map, the remaining operations are the same as those in the variable-based attention module.
After these operations, we can obtain output representation {\small$\bm{H}^{(tem)} \in \mathbb{R}^{T \times D}$} that captures the dynamic temporal dependency in bidding trajectories.

% \textbf{Remark:} The two attention mechanisms we designed can portray the bidding environment from variable and time perspectives. 
% This allows our model to fully exploit the dynamics behind the bidding environment to make accurate predictions. 
% This is the first approach to comprehensively model the bidding environment from different perspectives simultaneously. Compared to methods that model the environment from a single view, our approach can provide a more refined and complete delineation of the bidding environment. 

\subsubsection{\textbf{Variable-aware Fusion}} To make the model better understand the complex bidding environment, we fuse the representations from the variable and temporal perspectives. Since different variables describe the bidding environment from different views, we propose a Variable-aware Fusion (VaF) method to preserve environment diversity and enhance our model's robustness.

Specifically, given variable representation matrix {\small$\bm{H}^{(var)}$}, we extract the target variables' representations as {\small$\bm{H}_p^{(var)} \in \mathbb{R}^{C_2 \times D}$} and iteratively utilize each row to generate fused representation. 
Let {\small$\bm{h}_i^{(var)} \in \mathbb{R}^{D}$} be the representation of the $i$-th target variable. We fuse it with the temporal representation matrix {\small$\bm{H}^{(tem)} \in \mathbb{R}^{T \times D}$} via 
\begin{equation}\small
    \bm{H}_i = \sigma\left(\hat{\bm{H}}_i\right) \odot \bm{H}^{(tem)},
\end{equation}
where $\odot$ denotes the element-wise Hadamard product.
The sigmoid gate {\small$\sigma(\hat{\bm{H}}_i)$} controls which input {\small$\bm{H}^{(tem)}$} is relevant for forecasting future states of the $i$-th target variable. 
Its input {\small$\hat{\bm{H}}_i$} is produced by {\small$\hat{\bm{H}}_i = \mathrm{MLP}\left(\mathrm{Concat}\left(\bm{h}_i^{(var)}, \bm{H}^{(tem)}\right)\right),$}
% \begin{equation}
%     \hat{\bm{H}}_i = \mathrm{MLP}\left(\mathrm{Concat}\left(\bm{h}_i^{(var)}, \bm{H}^{(tem)}\right)\right),
% \end{equation}
where {\small$\mathrm{MLP}$} is a two-layer fully connected network, and {\small$\mathrm{Concat}$} represents the concatenation operation with broadcasting. 

\textbf{Remark:} 
% The two attention mechanisms we designed in the bidding transformer can depict bidding environments from variable and temporal perspectives, leading to a refined and complete delineation of their dynamics.
The two attention mechanisms we designed in the bidding transformer can depict bidding environments from the perspective of variables and time respectively, thus providing a fine and complete description of their dynamics.
Moreover, when the environment changes, variable attention can function as a context learner to perceive the new dependency between variables. This can enhance our model's ability to adapt to new environments, which is empirically verified in Section~\ref{sec:exp_main} and \ref{sec:ablation}.

\subsection{Zero-inflated Projection \& Auxiliary Task}\label{sec:zero_inflated}

\subsubsection{\textbf{Zero-inflated Projection}}
To model the unique zero-inflated distribution of bidding data, we enable our model to be aware of the zero information by a binary classifier and incorporate such information into the prediction of bidding results for joint optimization.

Specifically, given the embedding of the $i$-th target variable $\bm{H}_i$, we estimate the probability of the target value $y_i$ not be zero via
\begin{equation}\small\label{eq:zi_cls}
\mathrm{Pr}(y_i \neq 0) = \sigma(\mathrm{MLP}(\bm{H}_i)),
\end{equation}
where {\small$\sigma(\cdot)$} is the sigmoid function. 
On top of this, we produce the value prediction by incorporating the non-zero probability: 
\begin{equation}\small
    \hat{y}_i = \mathrm{Pr}(y_i \neq 0)\cdot \tilde{y}_i + (1-\mathrm{Pr}(y_i \neq 0)) \cdot 0, \label{eq:zeroinfloss}
\end{equation}
where {\small$\tilde{y}_i = \mathrm{MLP}(\bm{H}_i)$} is the prediction of the target value's magnitude. Since it takes into the zero-inflated phenomenon while projecting the target variable, we name it ``zero-inflated projection''.

Finally, we can derive a joint optimization objective that combines cross entropy (CE) loss and mean square error (MSE) loss:
\begin{equation}\small
\mathcal{L}^{(z)}_i = \underbrace{\mathrm{Pr}(y_i \neq 0)\cdot I_{\{y_i \neq 0\}} + (1-\mathrm{Pr}(y_i \neq 0))\cdot I_{\{y_i = 0\}}}_{\mathrm{CE}} + \underbrace{(y_i - \hat{y}_i)^2}_{\mathrm{MSE}},
\end{equation}
where $I_{\{\cdot\}}$ is the indicator function.
This joint objective allows the model predictions to converge to the zero-inflated distribution, with a theoretical guarantee provided in Section~\ref{sec:thm} and empirical evaluation in Appendix~\ref{appx:zip}.
By adding the loss function of all $C_2$ target variables, we obtain the overall loss for zero-inflated projection:
\begin{equation}\small\label{eq:loss_z}
    \mathcal{L}^{(z)} = \sum_{i=1}^{C_2}\mathcal{L}^{(z)}_i.
\end{equation}

% \textbf{Remark:} 
Existing studies for zero-inflated distribution usually take a two-step paradigm, where the first step trains a classifier to determine whether the target value is zero and the second trains a projector for only non-zero values~\cite{cheung2002zero}. 
The performance of these methods is limited by the classifier, which estimates a hard binary label for each sample. Consequently, if the classifier predicts non-zero samples with zero labels, the projector will be disabled, leading to large prediction errors.
In contrast, our model takes an end-to-end approach that estimates a soft label $\mathrm{Pr}(y_i \neq 0)$ for each sample. It can work together with the predictor to control the final predicted value. This allows the predictor to correct its forecast when the classifier's prediction is not up to standard. Additionally, this comprehensive approach streamlines the model training process.
% This end-to-end approach allows us to produce a joint optimization objective that includes both the classification and prediction loss functions. 
 
\subsubsection{\textbf{Cumulative Prediction}}
To give our model a global view of bidding environments, we propose a self-supervised auxiliary task that forecasts future cumulative information using the target variable representation, as in the upper left part of \fig~\ref{fig:framework}. 

Specifically, given the $i$-th target variable's representation $\bm{H}_i$, we predict the cumulative value from the current time slot to the end of the ad campaign of the target variable by
\begin{equation}\small
    \hat{y}^{(c)}_i = \mathrm{MLP}(\bm{H}_i),
\end{equation}
where {\small$\hat{y}^{(c)}_i$} is the prediction result. With its ground truth {\small$y^{(c)}_i$}, we can optimize this task by a mean square error loss:
\begin{equation}\small\label{eq:loss_c}
    \mathcal{L}^{(c)} = \sum_{i=1}^{C_2}\left(\hat{y}^{(c)}_i - y^{(c)}_i\right)^2.
\end{equation}

\subsubsection{\textbf{Model Training}}
Based on the zero-inflated projection loss in \equ~\eqref{eq:loss_z} and cumulative prediction loss in \equ~\eqref{eq:loss_c}, we can obtain the overall optimization objective as follows:
\begin{equation}\small
    \mathcal{L} = \mathcal{L}^{(z)} + \gamma \mathcal{L}^{(c)},
\end{equation}
where $\gamma$ is a balancing factor. 
With loss $\mathcal{L}$, our model can be trained by the backpropagation algorithm in an end-to-end manner.

\section{Theoretical Analysis}\label{sec:thm}

% \subsection{Proper Loss Theorem}

In order to study the theoretical properties of the proposed zero-inflated projection loss, we consider a more general supervised learning setting where we are interested in learning a mapping $f: \mathbb{R}^d \to \mathbb{R}$ given empirical data $\{\bm{x}, y\}_{i=1}^m$ with $\bm{x} \in \mathbb{R}^d$ and $y \in \mathbb{R}$. Similar to Eq.~\eqref{eq:zeroinfloss}, we define a zero-inflated model \citep{cheung2002zero} for $Y$, in which the expectation can be written as
\begin{equation}\label{eq:zinf}\small
    \mathbb{E} [Y|\bm{x}] \triangleq p(\bm{x}) \cdot g(\bm{x}) + (1 - p(\bm{x})) \cdot 0, 
\end{equation}
where $p$ is the true probability of $Y$ being in the non-negative group and $g$ is a deterministic function standing for the non-negative ground-truth expected label. We propose the following optimization objective:
\begin{equation*}\small
    \min_{\theta, \phi} \mathcal{L} \coloneqq \mathbb{E}_{(X, Y) \sim \tilde{\mathbb{P}}} (p_{\theta}(\bm{X})g_{\phi}(\bm{X}) - Y)^2 + H(p(\bm{X}) | p_{\theta}(\bm{X})),
\end{equation*}
where {\small$H(p | p_{\theta}) = -p \ln{p_{\theta}} - (1 - p) \ln{(1 - p_{\theta})}$} is the binary cross entropy between $p$ and $p_{\theta}$.

We next show that the learned conditional label distribution converges to the true distribution in the true-data setting.
\begin{theorem}[Zero-inflated Proper Loss]
Let $\mathcal{F}$ be a set of all measurable functions and $p_{\theta}, g_{\phi} \in \mathcal{F}$. Assume that $p$ and $g$ are conditionally independent, then, with respect to the zero-inflated model defined in Eq.~\eqref{eq:zinf}, we have $p_{\theta^*}(\bm{x}) = p(\bm{x})$ and $p_{\theta^*}(\bm{x})g_{\phi^*}(\bm{x}) = p(\bm{x})g(\bm{x})$, therefore $f(\bm{x}) = \mathbb{E} [Y|\bm{x}]$ for any $\bm{x} \in \mathbb{R}^d$. In addition, $g_{\phi^*}(\bm{x})$ is generally not unique.
\end{theorem}
To be brief, we show the consistency of the proposed loss by taking advantage of standard KKT conditions and implications of Hessian matrices with respect to convexity. We defer the detailed proof to Appendix~\ref{sec:proof}. 
% Empirical evaluation is in \fig~\ref{fig:zip} in Appendix~\ref{appx:zip}.

% Table generated by Excel2LaTeX from sheet '10main'
\begin{table*}[t]
  \centering
  \caption{Performance comparison over eight datasets in terms of MAE and RMSE \wrt three target variables: cost, reward, count. ``-'' means the model cannot produce corresponding predictions. The bold/underlined font means the best/the second-best result.}%\vspace{-2mm}
  \resizebox{\linewidth}{!}{
  \setlength{\tabcolsep}{0.6mm}
  % \setstretch{1.12} 
  % \renewcommand{\arraystretch}{1.4}
    \begin{tabular}{llcccccccccccccccc}
    \toprule
          & Dataset & \multicolumn{2}{c}{BCB} & \multicolumn{2}{c}{TR} & \multicolumn{2}{c}{BS} & \multicolumn{2}{c}{BM} & \multicolumn{2}{c}{BL} & \multicolumn{2}{c}{D6} & \multicolumn{2}{c}{D12} & \multicolumn{2}{c}{D18} \\
    \cmidrule(r){3-4} \cmidrule(r){5-6} \cmidrule(r){7-8} \cmidrule(r){9-10} \cmidrule(r){11-12} \cmidrule(r){13-14} \cmidrule(r){15-16} \cmidrule(r){17-18}
     & Metric & MAE   & RMSE  & MAE   & RMSE  & MAE   & RMSE  & MAE   & RMSE  & MAE   & RMSE  & MAE   & RMSE  & MAE   & RMSE  & MAE   & RMSE \\
    \midrule
    \multirow{10}[2]{*}{\rotatebox{90}{\textbf{Cost}}} & DLF   & 177.22  & 1038.73  & 337.56  & 2288.56  & 188.69  & 435.47  & 1317.44  & 2364.62  & 7130.59  & 13144.37  & 386.03  & 1225.56  & 266.31  & 1026.17  & 233.19  & 998.94  \\
          & ADM   & 164.26  & 992.28  & 279.40  & 1382.48  & 187.83  & 432.65  & 1222.74  & 2314.76  & 8492.74  & 15877.66  & 356.57  & 1168.13  & 249.72  & 907.07  & 201.67  & 837.37  \\
          & LP    & 521.28  & 5513.79  & 670.94  & 5652.17  & 630.71  & 5551.63  & 2881.14  & 14380.97  & 16479.33  & 44689.66  & 626.78  & 6085.45  & 470.77  & 4992.03  & 458.36  & 4823.93  \\
          & MMP   & 165.08  & \underline{792.29}  & 257.96  & 1195.95  & 199.92  & 471.19  & 1520.61  & 35308.12  & 11211.28 & 29587.71 & 218.11  & 971.93  & 151.22  & 803.43  & 131.16  & 724.20  \\
          & MLP   & \underline{157.60}  & 814.94  & 223.84  & \underline{914.96}  & 184.27  & \underline{429.32}  & \underline{975.87}  & \underline{2040.10}  & 8427.86  & 16257.88  & \underline{198.76}  & \underline{832.21}  & \underline{139.27}  & 892.18  & 119.63  & 647.71  \\
          & iTransformer & 183.83  & 949.73  & 269.15  & 1278.51  & 207.98  & 487.07  & 1223.86  & 2467.54  & 7708.04  & 14322.90  & 243.03  & 1130.71  & 165.90  & 938.27  & 133.25  & 744.00  \\
          & NST   & 161.79  & 965.66  & 234.83  & 1103.03  & 186.42  & 440.66  & 1057.50  & 2141.10  & 8222.52  & 14977.61  & 208.37  & 909.10  & 149.90  & 859.11  & 120.81  & 616.08  \\
          & Informer & 159.31  & 805.66  & 239.38  & 1141.87  & \underline{180.92}  & 450.02  & 1065.46  & 2220.36  & 7952.92  & 15233.91  & 209.18  & 944.01  & 145.15  & 803.10  & 120.96  & 664.25  \\
    \cmidrule{2-18}
          & Informer(fm) & 158.55  & 796.40  & \underline{214.56}  & 1060.32  & 187.06  & 448.36  & 1015.16  & 2115.96  & \underline{6037.38}  & \underline{12460.79}  & 206.43  & 944.93  & 140.57  & \underline{759.19}  & \underline{113.55}  & \underline{607.79}  \\
          & \textbf{Bid2X}(fm) & \textbf{122.50} & \textbf{589.35} & \textbf{172.40} & \textbf{792.90} & \textbf{148.28} & \textbf{370.69} & \textbf{751.83} & \textbf{1618.24} & \textbf{4170.21} & \textbf{8758.04} & \textbf{158.81} & \textbf{717.49} & \textbf{109.27} & \textbf{553.85} & \textbf{88.96} & \textbf{425.67} \\
    \midrule{}
    \multirow{10}[2]{*}{\rotatebox{90}{\textbf{Reward}}} & DLF   & 8.15  & 58.33  & 12.80  & 162.93  & 16.23  & 66.72  & 64.07  & 286.65  & 281.75  & 1249.32  & 39.61  & 302.77  & 24.68  & 177.88  & 17.43  & 149.91  \\
        & ADM   & 8.01  & 47.52  & 12.27  & 156.54  & 15.87  & 60.99  & 51.77  & 232.26  & 296.32  & 1366.17  & 30.95  & 273.44  & 21.25  & 145.96  & 14.21  & 124.81  \\
        & LP    & 1532.06  & 23559.05  & 2077.69  & 42414.41  & 2006.52  & 25790.00  & 8094.40  & 61671.17  & 43145.00  & 303338.10  & 1905.18  & 25388.42  & 1415.57  & 29026.07  & 1155.66  & 22983.07  \\
          & MMP   & 8.27  & 49.26  & 13.06  & 176.42  & 10.49  & 45.89  & 48.59  & 142.20  & 652.76  & 3698.89  & 10.97  & 62.07  & 7.54  & 99.03  & 5.71  & 43.97  \\
          & MLP   & 8.12  & 47.84  & \underline{10.98}  & \underline{149.19}  & 9.96  & 42.50  & 42.14  & \underline{120.07}  & 294.89  & 1368.50  & 9.93  & \underline{52.17}  & 7.13  & 61.81  & 4.91  & \underline{32.87}  \\
          & iTransformer & 8.06  & 49.01  & 12.24  & 155.90  & 10.14  & 45.28  & 46.89  & 136.52  & 254.78  & \underline{1124.58}  & 10.85  & 62.31  & 7.34  & 89.07  & 5.21  & 41.63  \\
          & NST   & 7.47  & 49.30  & 11.05  & 148.25  & 9.50  & \underline{41.65}  & 43.64  & 133.94  & 248.60  & 1156.04  & 10.63  & 305.18  & 7.45  & 271.76  & 4.97  & 38.05  \\
          & Informer & 7.46  & \underline{44.35}  & 11.64  & 170.88  & 9.55  & 42.63  & 42.98  & 127.11  & 295.86  & 1380.27  & 9.97  & 55.48  & 6.85  & 95.12  & 5.01  & 41.72  \\
    \cmidrule{2-18}
          & Informer(fm) & \underline{7.44}  & 44.42  & 11.24  & 162.57  & \underline{9.42}  & 42.20  & \underline{42.05}  & 125.00  & \underline{233.72}  & 1252.80  & \underline{9.92}  & 56.18  & \underline{6.74}  & \underline{91.36}  & \underline{4.82}  & 38.12  \\
          & \textbf{Bid2X}(fm) & \textbf{5.94} & \textbf{35.28} & \textbf{8.90} & \textbf{131.92} & \textbf{7.76} & \textbf{35.65} & \textbf{31.91} & \textbf{99.01} & \textbf{170.03} & \textbf{999.59} & \textbf{7.82} & \textbf{44.63} & \textbf{5.45} & \textbf{74.23} & \textbf{4.01} & \textbf{27.68} \\
    \midrule
    \multirow{10}[2]{*}{\rotatebox{90}{\textbf{Count}}} & DLF   & 22.04  & 104.71  & 28.73  & 144.89  & 52.10  & 167.94  & 122.49  & 326.37  & 475.95  & 1241.22  & 71.22  & 286.95  & 46.49  & 229.12  & 35.10  & 279.45  \\
        & ADM   & 20.46  & 98.49  & 24.93  & 108.11  & 49.09  & 152.89  & 88.12  & 291.09  & 510.54  & 1263.78  & 47.42  & 242.83  & 35.47  & 194.74  & 28.81  & 227.71  \\
        & LP    & -     & -     & -     & -     & -     & -     & -     & -     & -     & -     & -     & -     & -     & -     & -     & - \\
          & MMP   & 20.23  & 98.57  & 24.30  & 102.68  & 25.24  & 85.18  & 93.96  & 372.61  & 862.19  & 5439.32  & 23.90  & 114.03  & 17.99  & 87.49  & 17.41  & 86.01  \\
          & MLP   & \underline{18.54}  & \underline{92.12}  & 20.70  & 92.02  & \underline{22.74}  & \underline{78.46}  & 75.29  & 274.30  & 302.62  & 750.63  & 22.87  & 109.58  & \underline{16.26}  & \underline{78.44}  & 16.95  & 79.97  \\
          & iTransformer & 21.75  & 103.70  & 24.91  & 104.19  & 26.75  & 88.34  & 91.51  & 302.15  & 290.17  & 713.01  & 25.96  & 118.74  & 19.47  & 92.26  & 17.64  & 85.13  \\
          & NST   & 19.35  & 105.96  & 22.39  & 94.04  & 23.63  & 80.60  & 81.31  & 278.85  & 315.69  & 751.85  & 22.99  & \underline{107.69}  & 17.19  & 80.49  & 16.53  & 78.88  \\
          & Informer & 19.27  & 97.05  & 22.96  & 99.89  & 23.68  & 82.45  & 81.74  & 290.02  & 313.97  & 784.46  & 22.99  & 111.42  & 17.44  & 86.46  & 16.39  & 82.05  \\
    \cmidrule{2-18}
          & Informer(fm) & 19.12  & 95.93  & \underline{20.38}  & \underline{91.81}  & 23.42  & 81.63  & \underline{73.92}  & \underline{267.62}  & \underline{246.93}  & \underline{672.31}  & \underline{22.60}  & 109.88  & 16.82  & 83.00  & \underline{15.38}  & \underline{77.00}  \\
          & \textbf{Bid2X}(fm) & \textbf{12.02} & \textbf{72.07} & \textbf{13.26} & \textbf{66.96} & \textbf{14.54} & \textbf{58.85} & \textbf{47.21} & \textbf{210.57} & \textbf{151.70} & \textbf{475.96} & \textbf{14.30} & \textbf{84.57} & \textbf{10.53} & \textbf{61.03} & \textbf{9.23} & \textbf{50.73} \\
    \bottomrule
    \end{tabular}%
   }%\vspace{-2mm}
   \label{tab:main}%
\end{table*}%

\section{Experiments}\label{sec:exp}

\subsection{Experimental Settings}

\subsubsection{\textbf{Dataset \& Baseline}} To evaluate the performance of \model, we conduct extensive experiments on eight ad bidding datasets containing 132,734,218 bidding trajectories and 3,070,868 bidding records from the advertising platform in Taobao, one of the largest e-
commerce platforms worldwide.
The datasets cover multiple types of bidding strategies (\eg BCB~\cite{zhou2008budget}, TargetROAS~\cite{feng2023online}), spanning various advertisers with various budgets and delivery durations. \tab~\ref{tab:data_basic} and \tab~\ref{tab:data_stat} in Appendix~\ref{appx:exp_set} summarize the datasets, illustrating diversity across scenarios.

% We choose Mean Average Error (MAE) and Root Mean Square Error (RMSE) for performance evaluation. A lower metric value indicates better performance.
% We select 9 methods as baselines, which can be categorized into three distinct groups: 
% $i)$ \textbf{Heuristic method:} historical log-based LP~\cite{he2021unified}. 
% $ii)$ \textbf{Variable-based approaches:} MMP~\cite{liu2021neural} and MLP~\cite{guo2024aigb}, which model the variable relations at each time step independently. We also compare with iTransformer~\cite{liuitransformer} that models the variable relations using the whole bidding trajectory.
% $iii)$ \textbf{Temporal-based approaches:} non-stationary transformer (NST)~\cite{liu2022non} and Informer~\cite{zhou2021informer} . For a fair comparison, we also train Informer for all datasets in a foundation-model manner, denoted as Informer(fm).

% point estimator: Point estimators are limited to providing enough landscape information for advertisers
% series: 
% landscape forecasting:将骨干模型产生的表征拼接上MLP用于outcome预测
We choose Mean Average Error (MAE) and Root Mean Square Error (RMSE) for performance evaluation. A lower metric value indicates better performance.
The baselines we compare can be divided into four streams. 
$i)$ \textbf{Landscape forecasting approaches} model the market competition in online advertising by forecasting the probability distribution of market price for each ad auction.
The compared models includes DLF~\cite{ren2019deep} and ADM~\cite{li2022arbitrary}. We adapt them by feeding the hidden representations before the final prediction layers into an MLP for bid outcome prediction.
$ii)$ \textbf{Heuristic method} includes historical log-based LP~\cite{he2021unified}. 
$iii)$ \textbf{Variable-based approaches} aims to model the relations between bid and its outcomes, including MMP~\cite{liu2021neural} and MLP~\cite{guo2024aigb} that consider variable relations at bid record level, and iTransformer~\cite{liuitransformer} considering relation at the bidding trajectory level.
$iv)$ \textbf{Temporal-based approaches} mainly consider the dynamic sequence dependency of different bid records in a bidding trajectory. The compared baselines include non-stationary transformer (NST)~\cite{liu2022non} and Informer~\cite{zhou2021informer}. 
For a fair comparison, we also train Informer for all datasets in a foundation-model manner, denoted as Informer(fm).

% $i)$ \textbf{Point-based methods} directly model the relationship between bid and its outcomes within each bid record, including LP~\cite{he2021unified}, Min-Max Predictor (MMP)~\cite{liu2021neural} and a common environment model in model-based RL for auto-bidding (\ie MLP)~\cite{guo2024aigb}.
% $ii)$ \textbf{Series-based methods} consider the sequence dependency of different bid records and view a bidding trajectory as a whole. Baselines are mainly derived from recent time series forecasting models, including iTransformer~\cite{liuitransformer}, non-stationary transformer (NST)~\cite{liu2022non} and Informer~\cite{zhou2021informer}. For fair comparison, we also train Informer for all datasets in a foundation-model manner, denoted as Informer(fm).

\subsubsection{\textbf{Implementation Protocol}}
Our \model is implemented with PyTorch 2.2.0 with an NVIDIA L40S GPU. The number of layers is set to 2. The hidden dimension $D$ is searched from 32 to 1024. For the task balancing factor $\gamma$, we test it from $\{0.5, 0.75, 1, 1.25, 1.5\}$. Our model is trained using Adam optimizer with a learning rate of 0.0001 and batch size of 128. The final parameter setting of our model is according to \fig~\ref{fig:para} in Appendix~\ref{appx:param_sens}. 

\subsection{Overall Performance}\label{sec:exp_main}

We split every dataset into training, validation, and test sets in a ratio of 1:1:1. For baselines, we train a dedicated model for each dataset, while \model is trained and evaluated across all datasets in a one-for-all foundation paradigm.
\tab~\ref{tab:main} presents the comparison results of the environment modeling task. We have the following key findings: 
\textbf{(1)} \model outperforms all competing baselines across every target variable on eight datasets with a large margin.
 % (??\% and ??\% reduction in MAE and RMSE)
This underscores the significant potential and benefits of our bidding foundation model where different datasets can benefit each other for learning generalized bidding laws.
\textbf{(2)} Baseline approaches exhibit inconsistent performance across diverse datasets, indicating their instability across scenarios. In contrast, our model consistently offers more stable and reliable results, highlighting its robustness and adaptability to various bidding scenarios. 
\textbf{(3)} Interestingly, Informer(fm) performs worse than baselines dedicated for each dataset in some cases, suggesting that Informer struggles to adapt to these distinct data distributions relying solely on temporal dependency modeling.
% \textbf{(4)} Approaches that focus on modeling the current dynamic environment, whether variable-based or temporal-based models, generally outperform heuristic ones that rely on a stationary assumption between the current and historical environment and only utilize historical logs for prediction. This reflects the importance of capturing environmental dynamics from both variable and temporal perspectives.
\textbf{(4)} Approaches that focus on modeling the dynamic environment, whether variable or temporal-based models, generally outperform heuristic methods that rely on a stationary assumption of the current and historical environment. This reflects the importance of capturing environmental dynamics from both variable and temporal perspectives.
% Landscape forecasting methods does not generate that used for market price estimation.

% 此外，忽略独特的零膨胀数据分布也是informer(fm)难以准确建模出价环境的原因，具体分析见xx。
% In addition, ignoring the unique zero-inflated data distribution is also the reason why informer (fm) has difficulty in accurately modeling the bidding environment. See xx for detailed analysis.

\subsection{Zero-Shot Performance}

\begin{figure}
    \centering
    \includegraphics[width=0.95\columnwidth]{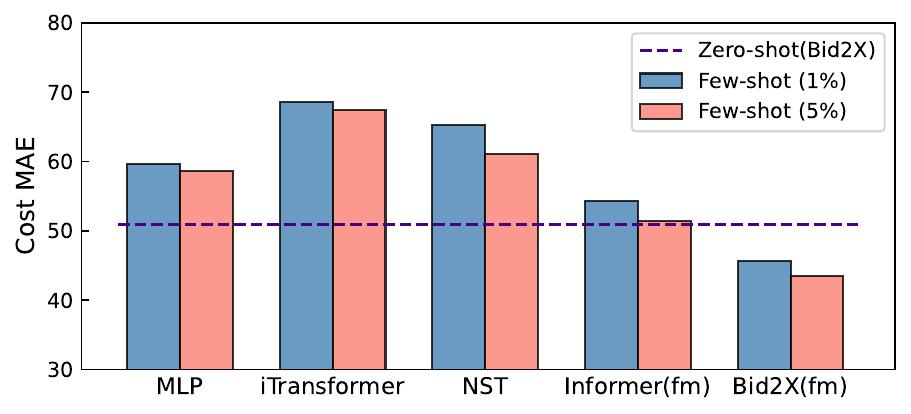}\vspace{-2mm}
    \caption{Few-shot performance of \model and baselines on the crowd-type bidding scenario with 1\% and 5\% training data. Zero-shot results of our \model are given by the dashed line.}%\vspace{-5mm}
    \label{fig:zero_shot}
\end{figure}

The remarkable feature of the foundation model is its excellent generalizability. Few-shot and zero-shot generalizations are often used as the ultimate task to assess this ability.
To this end, we evaluate our Bid2X through the few-shot learning and zero-shot generalization in this part.
Specifically, we select a new dataset unseen by any model in \tab~\ref{tab:main} from the crowd-type bidding scenario.
We split it into training, validation, and test sets in a ratio of 1:1:1. In few-shot learning, only a restricted amount of training data, namely 1\% and 5\%, are utilized for model learning. We compare our \model with some competitive baselines.
To increase the task difficulty, we also evaluate the zero-shot performance of \model via direct testing without a training phase.
\tab~\ref{tab:few_shot} in Appendix~\ref{appx:few_shot} provides the full few-shot results. Due to limited space, we visualize the MAE result regarding the ``cost'' target variable in \fig~\ref{fig:zero_shot}.
As observed, \model demonstrates a superior generalization performance over the baseline models. 
This generalization can be attributed to the successful transfer of learned variable relationships. 
This suggests general rules among different advertising bidding scenarios, which is worthy of further exploration by the community.

\begin{table*}[t]
  \centering
  \caption{Ablation study on key components of our model in terms of MAE with the best in bold. Rwd: Reward. Cnt: Count.}%\vspace{-.2cm}
  \resizebox{\linewidth}{!}{
  \setlength{\tabcolsep}{0.4mm}
  % \setstretch{1.12} 
  \renewcommand{\arraystretch}{1.15}
    \begin{tabular}{lcccccccccccccccccccccccc}
    \toprule
    Dataset & \multicolumn{3}{c}{BCB} & \multicolumn{3}{c}{TR} & \multicolumn{3}{c}{BS} & \multicolumn{3}{c}{BM} & \multicolumn{3}{c}{BL} & \multicolumn{3}{c}{D6} & \multicolumn{3}{c}{D12} & \multicolumn{3}{c}{D18} \\
    \cmidrule(r){2-4} \cmidrule(r){5-7} \cmidrule(r){8-10} \cmidrule(r){11-13} \cmidrule(r){14-16} \cmidrule(r){17-19} \cmidrule(r){20-22} \cmidrule(r){23-25}
    Target & Cost  & Rwd & Cnt & Cost  & Rwd & Cnt & Cost  & Rwd & Cnt & Cost  & Rwd & Cnt & Cost  & Rwd & Cnt & Cost  & Rwd & Cnt & Cost  & Rwd & Cnt & Cost  & Rwd & Cnt \\
    \midrule
    \textbf{Bid2X} & \textbf{122.50} & \textbf{5.94} & \textbf{12.02} & \textbf{172.40} & \textbf{8.90} & \textbf{13.26} & \textbf{148.28} & \textbf{7.76} & \textbf{14.54} & \textbf{751.83} & \textbf{31.91} & \textbf{47.21} & \textbf{4170.21} & \textbf{170.03} & \textbf{151.70} & \textbf{158.81} & \textbf{7.82} & \textbf{14.30} & \textbf{109.27} & \textbf{5.45} & \textbf{10.53} & \textbf{88.96} & \textbf{4.01} & \textbf{9.23} \\
    r/p va    & 135.06  & 6.39  & 12.73  & 195.39  & 9.86  & 14.28  & 151.91  & 8.08  & 15.27  & 885.30  & 36.13  & 53.23  & 6185.36  & 226.93  & 186.18  & 175.85  & 8.47  & 15.24  & 120.96  & 5.90  & 11.16  & 100.74  & 4.31  & 9.81  \\
    w/o va    & 150.28  & 6.34  & 13.58  & 219.35  & 9.17  & 15.44  & 169.93  & 8.19  & 16.67  & 1031.83  & 35.37  & 57.99  & 6577.34  & 187.51  & 193.17  & 194.41  & 8.34  & 16.13  & 136.83  & 5.72  & 12.06  & 116.66  & 4.35  & 10.91  \\
    w/o ta    & 159.00  & 7.16  & 17.61  & 227.74  & 10.24  & 19.60  & 178.12  & 9.11  & 21.43  & 1022.72  & 39.94  & 71.88  & 8026.04  & 208.97  & 243.45  & 202.42  & 9.43  & 20.65  & 144.70  & 6.45  & 15.64  & 126.72  & 4.80  & 14.50  \\
    w/o zip    & 138.63  & 8.93  & 14.09  & 198.54  & 12.33  & 15.98  & 160.12  & 11.12  & 16.67  & 900.61  & 44.40  & 60.33  & 5445.56  & 236.57  & 238.45  & 179.49  & 11.61  & 16.86  & 122.67  & 7.63  & 12.19  & 104.91  & 6.18  & 11.00  \\
    w/o cfp    & 131.14  & 6.11  & 12.13  & 188.52  & 9.66  & 13.41  & 150.87  & 7.83  & 14.62  & 846.20  & 33.39  & 49.08  & 5574.34  & 228.87  & 161.43  & 168.36  & 8.03  & 14.39  & 118.03  & 5.76  & 10.64  & 102.27  & 4.18  & 9.46  \\
    \bottomrule
    \end{tabular}%
  }%\vspace{-2mm}
  \label{tab:ablation}%
\end{table*}%

\subsection{Further Analysis of \model}

\subsubsection{\textbf{Ablation Study}}\label{sec:ablation}
To verify our model design, we carry out ablation experiments on the following variants: 
(a) \textbf{r/p va} replaces the variable attention with temporal attention. 
(b) \textbf{w/o va} removes the variable attention encoder. 
(c) \textbf{w/o ta} disables the temporal attention modeling by padding the target entries of the decoder input as zeros. 
(d) \textbf{w/o zip} removes the zero-inflated projection by disabling the classification-related parts.
(e) \textbf{w/o cfp} does not use the cumulative future prediction task in \equ~\eqref{eq:loss_c}.
The MAE results of all datasets are shown in \tab~\ref{tab:ablation}. We can observe that all components contribute to the model’s overall performance.
Specifically, variants \textbf{w/o va} and \textbf{w/o ta} show large decreases, suggesting that our proposed variable and temporal attention are indispensable for effectively and comprehensively modeling the bidding environment.
Besides, the impact of removing these components on performance is more pronounced for the BL dataset compared to other datasets as its data possesses more complex relations and is more sensitive to environment modeling.

\begin{figure}[t]
    \centering
    \includegraphics[width=\columnwidth]{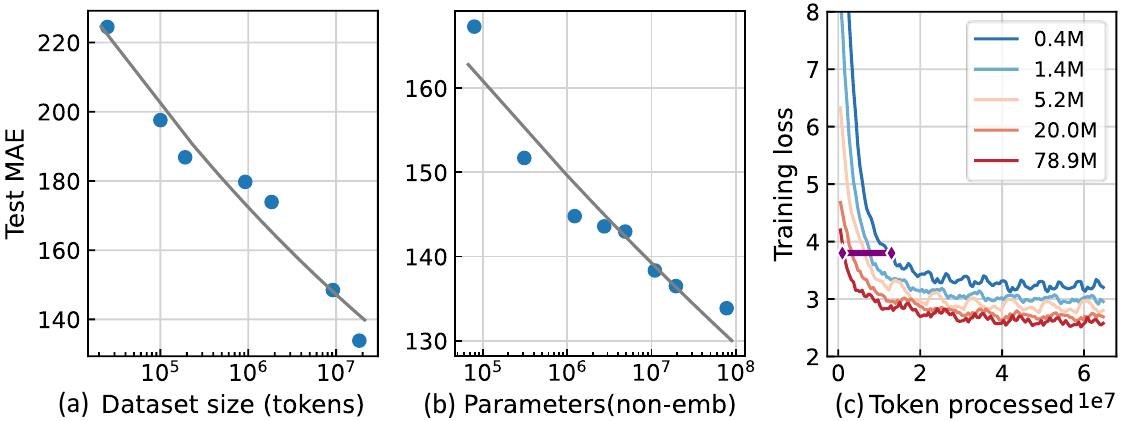}%\vspace{-2mm}
    \caption{Bidding environment modeling performance improves smoothly as we increase the (a) dataset size and (b) model size. (c) Training loss curve of varying-size models.}\vspace{-3mm}
    \label{fig:scale}
\end{figure}

% \begin{figure*}[t]
%     \centering
%     \subfigure[Scalability with data]{\includegraphics[width=0.3\columnwidth]{figures/scaling_law_data.pdf}}
%     \subfigure[Scalability with parameters]{\includegraphics[width=0.195\linewidth]{figures/scaling_law_param.pdf}}
%     \subfigure[Sample efficiency]{\includegraphics[width=0.18\linewidth]{figures/sample_eff.pdf}}
%     \subfigure[Monotonicity]{\includegraphics[width=0.185\linewidth]{figures/monotonicity.pdf}}
%     \subfigure[Predictability]{\includegraphics[width=0.19\linewidth]{figures/predictability.pdf}}
%     \caption{The power-law scalability with dataset size and parameters are shown in (a) and (b). The sample efficiency for different-size models is in (c). Our model's adherence to physics laws including monotonicity and predictability is in (d) and (e).}\label{fig:scale}%\vspace{-4mm}
% \end{figure*}

\subsubsection{\textbf{Scalability}}\label{sec:scale}

\begin{table}[b]
  \centering
  \caption{Monotonic ratio \wrt different cost levels.}%\vspace{-.2cm}
  \resizebox{0.9\linewidth}{!}{
    \begin{tabular}{lcccc}
    \toprule
    Tick cost & (0, 0.1k] & (0.1k, 1k] & (1k, 10k] & (10k, inf) \\
    \midrule
    Informer(fm) & 22.42\% & 22.26\% & 22.50\% & 25.66\% \\
    Bid2X(fm) & 82.36\% & 84.01\% & 85.44\% & 84.28\% \\
    \bottomrule
    \end{tabular}%
    }%\vspace{-3mm}
    \label{tab:mono_rate}%
\end{table}%

Scalability is a crucial characteristic of foundation models, therefore, we explore the scaling behavior of our \model \wrt dataset size $D$ and model size $N$. 
As shown in \fig~\ref{fig:scale}(a) and (b), we observe that the model performance improves predictably as the increase of $N$ and $D$, with trends spanning more than four orders of magnitude.
Specifically, model performance $L$ has a power-law relationship with each of the two scale factors $N$ ($L = (N / 3.87 \cdot 10^{75})^{-0.031}$) and $D$ ($L = (D / 3.81 \cdot 10^{38})^{-0.069}$).
Moreover, as in \fig~\ref{fig:scale}(c), increasing the model parameter size accelerates the convergence of the training loss. 
The purple horizontal line with diamond-shaped ends demonstrates that large models have higher sample efficiency than small models and achieve the same level of performance with fewer processed tokens.
These observations indicate that \model has shown scalability behaviors, wherein larger models generally exhibit improved performance.

\begin{figure}[t]
    \centering
    \subfigure[Monotonicity]{\includegraphics[width=0.48\columnwidth]{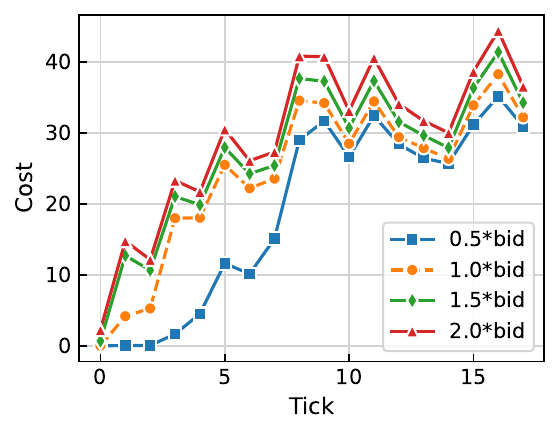}\label{fig:mono}}\vspace{-2mm}
    \subfigure[Predictability]{\includegraphics[width=0.49\columnwidth]{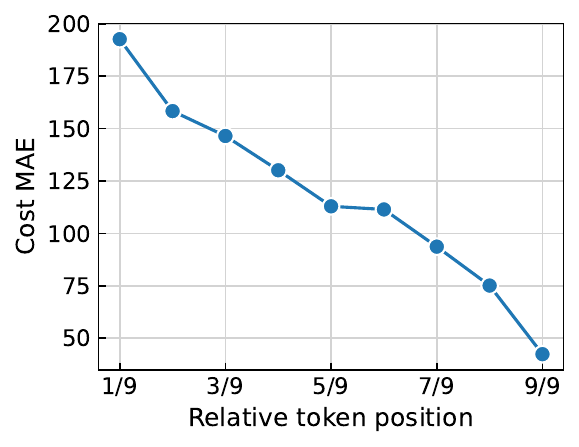}\label{fig:pred}}\vspace{-2mm}
    \caption{Our model's adherence to physics laws.}\vspace{-3mm}
    \label{fig:phy}
\end{figure}

\subsubsection{\textbf{Adherence to Physics Laws}}\label{sec:phy}

Some physical laws in ad bidding scenarios are the cornerstones of the model's ability to generalize across different scenarios, such as monotonicity and predictability. 
Monotonicity refers to the fact that for a given time step of an ad campaign, the tick cost monotonically increases with the bid. 
Predictability refers to the fact that for a given ad campaign, the more historical bidding information is available for that day, the more predictable the next time slot's results will be.

Firstly, we construct counterfactual samples to observe the monotonicity of the output cost by manually modifying its input control variable bid. 
Specifically, for each sample, we multiply its input bid by $\alpha$, with $\alpha$ belonging to $\{0.5,1.0,1.5,2.0\}$, and count the corresponding output cost during model inference.
If this set of cost is monotonically increasing with the bid, then we increase the hit number by one, otherwise the missing number is increased by one.
Finally, we calculate the monotonic ratio as the hit number divided by the total number, and the results are shown in \tab~\ref{tab:mono_rate}.
As observed, the monotonic ratio of our \model is significantly better than Informer(fm), indicating that our model truly captures the generalized law of monotonicity through diverse training data. 
In addition, we randomly select an ad campaign and visualize its monotonic output. As shown in \fig~\ref{fig:mono}, we can see that monotonicity is satisfied throughout the execution of the ad campaign. Interestingly, the lower bid has a greater impact on cost than higher bids, indicating that the cost of obtaining ad impressions does not increase linearly.

For predictability, since bidding trajectories have different lengths, we sample 9 time slots at medium intervals for each trajectory to calculate the prediction error.
The results are shown in \fig~\ref{fig:pred}.
It can be found that the prediction error decreases with the increase of the historical bidding trajectory length, which indicates that our model adheres to the predictability law. 
Meanwhile, this also verifies the validity of the sequence information that we model by temporal attention.

% \vspace{-1mm}
\subsection{Online A/B Testing Performance}

Besides offline evaluation, we deploy our model on a real-world online advertising environment in Taobao, one of the largest e-commerce platforms worldwide. The platform adjusts the bid in a real-time auction system for each incoming request based on marketing values and multiple constraints imposed by advertisers. Specifically, we re-utilize the learned foundation model and settle on the bid at the $(t+1)$-th time step corresponding to cost prediction as close to the remaining budget as possible: $b^* \in \arg\min_{b} |\sum_{t}c_{t}(b) - \text{budget}|$. Around one million ad campaigns set up by advertisers have been sampled for the experiment, which is a relatively small portion among all campaigns such that the auction environment is nearly stationary to any chosen campaign. We compare the adapted bidding model against model-based RL (MBRL), one of our top-performing bidding agents on the platform, in terms of the following metrics: number of page views (PV), consumed budget (Cost), the number of won impression opportunities in the episode (BuyCnt), gross merchandise value (GMV) and return on investment (ROI). As shown in \tab~\ref{tab:online_ab}, our model outperforms MBRL regarding the total consumed budget as well as all other metrics, which include the goal of maximizing GMV for advertisers, with only a slight increase in latency due to larger models. These improvements exhibit the effectiveness of \model given the fact that the deployed bidding model requires little adjustment to the foundation model.

\begin{table}[t]
  \centering
  \caption{Online A/B testing results spanning over $2$ months.}%\vspace{-3mm}
  \resizebox{\linewidth}{!}{
    \begin{tabular}{lcccccc}
    \toprule
     & PV & Cost & BuyCnt & GMV & ROI & Latency \\
    \midrule
    Bid2X vs MBRL & +7.57\% & +2.16\% & +3.89\% & +4.65\% & +2.44\% & +0.05 seconds \\
    \bottomrule
    \end{tabular}%
    }%\vspace{-5mm}
    \label{tab:online_ab}%
\end{table}%

\vspace{-1mm}
\section{Conclusion and Future Work}\label{sec:con}

This paper investigated the bidding environment modeling problem in online advertising by proposing a foundation model coined \model. 
Specifically, we encoded heterogeneous bidding data as series embeddings and jointly captured the complex and dynamic relationships via variable and temporal attention mechanisms.
We then devised a theoretically guaranteed zero-inflated projection module that combines non-zero probability estimation and value magnitude prediction to ensure the predicted value coverage of the true zero-inflated distribution.
Comprehensive experiments on eight bidding datasets and online A/B tests demonstrated the generalization of \model.
The future work lies in (1) further improving our proposed model's adherence to the laws of physics; (2) extending our \model to more bidding-related scenarios, \eg campaign effectiveness prediction. 

\bibliographystyle{ACM-Reference-Format}
\balance % balance the last page (2 even length columns)
\bibliography{5-sample-base}

% \newpage
\appendix

\section{Technical Proofs}\label{sec:proof}

\setcounter{theorem}{2}

\begin{theorem}[Zero-inflated Proper Loss]
Let $\mathcal{F}$ be a set of all measurable functions and $p_{\theta}, g_{\phi} \in \mathcal{F}$. Assume that $p$ and $g$ are conditionally independent, then, with respect to the zero-inflated model defined in Eq.~\eqref{eq:zinf}, we have $p_{\theta^*}(\bm{x}) = p(\bm{x})$ and $p_{\theta^*}(\bm{x})g_{\phi^*}(\bm{x}) = p(\bm{x})g(\bm{x})$, therefore $f(\bm{x}) = \mathbb{E} [Y|\bm{x}]$ for any $\bm{x} \in \mathbb{R}^d$. In addition, $g_{\phi^*}(\bm{x})$ is generally not unique.
\end{theorem}

\begin{proof}
Since $p$ and $g$ are conditionally independent, by taking the expectation over $Y$, the objective can be rewritten as
\begin{align*}%\small
    \min_{\theta, \phi} \mathbb{E}_{X \sim \tilde{\mathbb{P}}} (p_{\theta}(\bm{X})g_{\phi}(\bm{X}) - p(\bm{X})g(\bm{X}))^2 + H(p(\bm{X}) | p_{\theta}(\bm{X})).
\end{align*}
Furthermore, based on the i.~i.~d.~assumption, it is straightforward to focus on solving the sub-problem for each $\bm{X}$ independently:
\begin{align*}%\small
    \min_{\theta, \phi} \ell \coloneqq (p_{\theta}g_{\phi} - pg)^2 + H(p | p_{\theta}),
\end{align*}
where we omit the dependency on $X$ for ease of notation.

The stationarity conditions of $p_{\theta}$ and $g_{\phi}$ suggest that
\begin{align*}%\small
    & \frac{\partial}{\partial p_{\theta}} \ell = 2g_{\phi}(p_{\theta}g_{\phi} - pg) - \frac{p}{p_{\theta}} + \frac{1 - p}{1 - p_{\theta}} = 0 \\
    & \frac{\partial}{\partial g_{\phi}} \ell = 2p_{\theta}(p_{\theta}g_{\phi} - pg) = 0,
\end{align*}
which yields the solution $p_{\theta^*} = p$ and $p_{\theta^*} g_{\phi^*} = pg$ for $p_{\theta^*} \in (0, 1)$. This is applicable to the cases where $p_{\theta} \in \{0, 1\}$ when $p \in \{0, 1\}$ respectively because otherwise the entropy term  would be $\infty$. Thereafter, the optimality conditions hold for $p_{\theta^*} \in [0, 1]$. According to the assumption that $p_{\theta}$ and $g_{\phi}$ could be any measurable functions, the optimality conditions are satisfiable.

However, optimizing $\mathcal{L}$ does not guarantee a unique $g_{\phi^*}$ due to its non-convexity. This conclusion follows directly from the observation of its Hessian matrix:
\begin{align*}%\small
    \bm{H}_{\ell} \coloneqq
    \begin{bmatrix}
        2g_{\phi}^2 + \frac{p}{p_{\theta}^2} + \frac{(1 - p_{\theta})}{(1 - p_{\theta})^2} & 4 p_{\theta} g_{\phi} - 2pg  \\
        4 p_{\theta} g_{\phi} - 2pg & 2 p_{\theta}^2
    \end{bmatrix}.
\end{align*}
We know that $p_{\theta^*} = p$. Let $g_{\phi} = a g$ for some $a \in \mathbb{R}$. Furthermore, let $a > \frac{2}{3} + \sqrt{\frac{1}{9} + \frac{1}{6pg^2(1-p)}}$. The determinant therefore satisfies
\begin{align*}%\small
    \bm{H}_{\ell} = \frac{p}{1 - p} [(-12a^2+16a-4)pg^2(1-p)+2] < 0.
\end{align*}
The objective is thus non-convex for $p_{\theta}, g_{\theta} \in [0, 1]$.
\end{proof}

\section{Experimental Settings}\label{appx:exp_set}

% Table generated by Excel2LaTeX from sheet 'dataset'
\begin{table*}
  \centering
  \caption{Basic information of all datasets.}%\vspace{-2mm}
    \begin{tabular}{llll}
    \toprule
    Dataset & Description & \# Records & \# Trajectories \\
    \midrule
    BCB   & Data in budget constrained bidding scenarios & 47,879,102 & 1,128,142 \\
    TR    & Data in target ROAS scenarios & 7,601,466 & 187,030 \\
    BS    & Data of small-budget advertisers with the budget in (0, 5000] & 21,268,728 & 521,240 \\
    BM    & Data of middle-budget advertisers with the budget in (5000, 50000] & 3,353,924 & 78,426 \\
    BL    & Data of large-budget advertisers with the budget (50000, $\inf$) & 136,520 & 2,934 \\
    D6    & Data with ad delivery duration of 6-12 hours & 22,308,744 & 643,386 \\
    D12   & Data with ad delivery duration of 12-18 hours & 23,541,560 & 425,948 \\
    D18   & Data with ad delivery duration of 18-24 hours & 6,644,174 & 83,762 \\
    \bottomrule
    \end{tabular}%
  \label{tab:data_basic}%
\end{table*}%

\subsection{Datasets}

Here we provide more details of the datasets used in our study. We collect various bidding data from multiple scenarios and advertiser types. \tab~\ref{tab:data_basic} summarizes the basic information of the used datasets, and \tab~\ref{tab:data_stat} reports the basic statistics.  

% Table generated by Excel2LaTeX from sheet 'dataset'
\begin{table}[!h]
  \centering
  \caption{Statistics of datasets.}%\vspace{-2mm}
  \resizebox{\columnwidth}{!}{
  \setlength{\tabcolsep}{0.8mm}
  % \setstretch{1.12} 
  % \renewcommand{\arraystretch}{1.4}
    \begin{tabular}{lcccccccc}
    \toprule
          & \multicolumn{2}{c}{Bid} & \multicolumn{2}{c}{Cost} & \multicolumn{2}{c}{Reward} & \multicolumn{2}{c}{Count} \\
    \cmidrule(r){2-3} \cmidrule(r){4-5} \cmidrule(r){6-7} \cmidrule(r){8-9}
    Dataset & Mean  & SD    & Mean  & SD    & Mean  & SD    & Mean  & SD \\
    \midrule
    BCB   & 216.18  & 468.37  & 212.92  & 1068.67  & 10.00  & 65.20  & 25.15  & 101.25  \\
    TR    & 153.29  & 150.90  & 314.47  & 1417.75  & 15.42  & 180.82  & 28.85  & 103.65  \\
    BS    & 96.17  & 162.71  & 233.58  & 491.87  & 12.08  & 58.22  & 30.71  & 85.41  \\
    BM    & 172.38  & 288.93  & 1423.41  & 2475.33  & 61.95  & 221.72  & 104.97  & 278.37  \\
    BL    & 331.16  & 443.08  & 9800.88  & 13929.62  & 351.39  & 1148.51  & 372.36  & 689.95  \\
    D6    & 174.80  & 161.20  & 281.18  & 1307.49  & 13.48  & 87.47  & 29.60  & 116.15  \\
    D12   & 184.66  & 278.02  & 181.47  & 945.31  & 8.68  & 95.23  & 22.17  & 86.20  \\
    D18   & 220.97  & 331.04  & 140.56  & 762.13  & 5.77  & 38.36  & 19.14  & 77.93  \\
    \bottomrule
    \end{tabular}%
  }
  \label{tab:data_stat}%
\end{table}%

\subsection{Baselines}

We compare our \model with 9 baselines as follows:
\begin{itemize}[leftmargin=*]
    \item \textbf{DLF}~\cite{ren2019deep}: is a classic landscape forecasting model utilizing RNN-based approach.
    \item \textbf{ADM}~\cite{li2022arbitrary}: is a recent landscape forecasting literature, which devises a novel neighborhood likelihood loss for arbitrary distribution modeling.
    \item \textbf{LP}~\cite{he2021unified}: a heuristic method that utilizes linear programming to solve the environment by using historical bidding records, which may not match the current bidding environment due to its dynamic property.
    \item \textbf{MMP}~\cite{liu2021neural}: a min-max predictor that can ensure the partial monotonicity between the control variables and target variables, which is a widely-used model in bidding scenarios.  
    \item \textbf{MLP}~\cite{guo2024aigb}: a common environment model in model-based reinforcement learning for auto-bidding. It can directly learn the relations between control variables and target variables, but it neglects the sequence information inherent in bidding trajectories.
    \item \textbf{iTransformer}~\cite{liuitransformer}: a variable-based approach that takes each variable's series as a token and feeds it into a standard Transformer model. Since today's bidding data is not complete, we use historical bidding data that is complete as input.
    \item \textbf{non-stationary transformer (NST)}~\cite{liu2022non}:  a temporal-based series prediction approach aiming for the inherent non-stationarity problem, which is very common in bidding data. 
    \item \textbf{Informer}~\cite{zhou2021informer}: a temporal-based transformer model tailored for long-term series forecasting.
    \item \textbf{Informer(fm)}: we train Informer in a one-for-all foundation paradigm for fair comparison.
\end{itemize}

All baselines are separately trained on each dataset to learn a dedicated model except Informer(fm). The baseline Informer(fm) is trained on all datasets similar to our \model.

\section{Few-shot Performance}\label{appx:few_shot}

\tab~\ref{tab:few_shot} reports the few-shot learning results.
\begin{table}[h]
  \centering
  \caption{Few-shot performance of \model and baselines \wrt MAE.  The bold/underlined font means the best/the second-best result.}%\vspace{-2mm}
  \resizebox{\linewidth}{!}{
    \begin{tabular}{lcccccc}
    \toprule
    Task  & \multicolumn{3}{c}{Few-shot (1\%)} & \multicolumn{3}{c}{Few-shot (5\%)} \\
    \cmidrule(r){2-4} \cmidrule(r){5-7}
    Target & Cost  & Reward & Count & Cost  & Reward & Count \\
    \midrule
    MLP   & 59.66  & 3.59  & 58.58  & 58.68  & 3.40  & 58.29  \\
    iTransformer & 68.64  & 3.48  & 64.06  & 67.48  & 3.50  & 65.01  \\
    NST   & 65.25  & 3.50  & 61.95  & 61.08  & 3.29  & 56.69  \\
    \midrule
    Informer(fm) & \underline{54.34}  & \underline{3.20}  & \underline{56.83}  & \underline{51.41}  & \underline{3.18}  & \underline{55.80}  \\
    Bid2X(fm) & \textbf{45.63} & \textbf{2.65} & \textbf{38.09} & \textbf{43.50} & \textbf{2.55} & \textbf{36.76} \\
    \bottomrule
    \end{tabular}%
    }
  \label{tab:few_shot}%
\end{table}%

\section{Impact of Hyperparameters}\label{appx:param_sens}
% Table generated by Excel2LaTeX from sheet '10zero_shot'

\begin{figure}[t]
    \centering
    \includegraphics[width=\columnwidth]{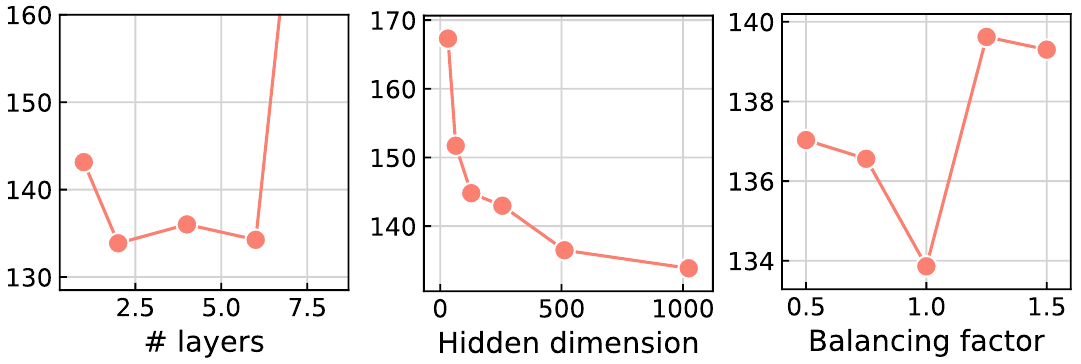}
    %\vspace{-2mm}
    \caption{Parameter sensitivity of \model using MAE metric.}%\vspace{-3mm}
    \label{fig:para}
\end{figure}

In this part, we conduct experiments to analyze the impacts of
critical hyperparameters: the number of layers, the hidden dimension, and the balancing factor.
In \fig~\ref{fig:para}, we present the MAE results on the mixed dataset with different parameters. 

Firstly, we keep the encoder and decoder layers the same and vary them in the set $\{1, 2, 4, 6, 8\}$. The results indicate that $2$ and $6$ deliver similar results. Considering the computation cost, we choose $2$ as the final setting.
Secondly, the hidden dimension is adjusted in the set $\{32, 64, 128, 256, 512, 1024\}$. The results indicate $1024$ as the optimal setting. Although the performance can increase with a larger dimensionality, we use $1024$ since it can produce satisfactory performance.
Thirdly, the effect of the balancing factor is tested in $\{0.5, 0.75, 1.0, 1.25, 1.5\}$. We can observe that a setting of 1.0 is optimal for our model.

\section{Distribution Visualization}\label{appx:zip}

\begin{figure}[b]
    \centering
    \includegraphics[width=\columnwidth]{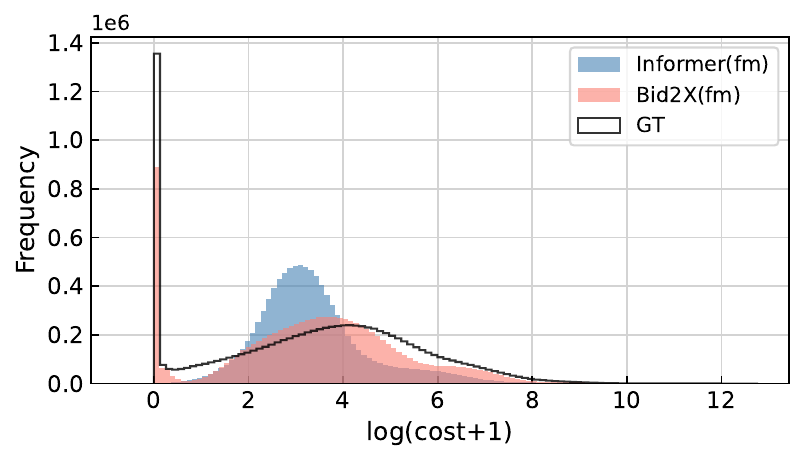}%\vspace{-2mm}
    \caption{Distribution of cost prediction. GT: Ground Truth.}%\vspace{-3mm}
    \label{fig:zip}
\end{figure}

In this part, we inspect the distribution of our model's prediction, to verify that our proposed zero-inflated projection can ensure the predicted data converges to the zero-inflated distribution.
We visualized the learned distribution for the cost variable in \fig~\ref{fig:zip}.
The results demonstrate that the cost predicted by our \model is consistent with the zero-inflated distribution. 
However, the baseline Informer that uses the classic neural network objective function converges to a normal distribution, which is inconsistent with the true bidding data distribution.

\end{document}